\documentclass[11pt,a4paper]{article}
\usepackage[hyperref]{emnlp2020}
\usepackage{times}
\usepackage{latexsym}

\usepackage{microtype}
\usepackage{tcolorbox}

\aclfinalcopy 

\title{When in Doubt, Ask: Generating Answerable and Unanswerable Questions, Unsupervised}

\author{Liubov Nikolenko\\
  Stanford University / Stanford, CA \\
  \texttt{liubov@stanford.edu} \\\And
  
  Pouya Rezazadeh Kalehbasti\\
  Stanford University / Stanford, CA \\
  \texttt{pouyar@stanford.edu} \\}

\date{}

\begin{document}
\maketitle
\begin{abstract}
Question Answering (QA) is key for making possible a robust communication between human and machine. Modern language models used for QA have surpassed the human-performance in several essential tasks; however, these models require large amounts of human-generated training data which are costly and time-consuming to create. This paper studies augmenting human-made datasets with synthetic data as a way of surmounting this problem. A state-of-the-art model based on deep transformers is used to inspect the impact of using synthetic answerable and unanswerable questions to complement a well-known human-made dataset. The results indicate a tangible improvement in the performance of the language model (measured in terms of F1 and EM scores) trained on the mixed dataset. Specifically, unanswerable question-answers prove more effective in boosting the model: the F1 score gain from adding to the original dataset the answerable, unanswerable, and combined question-answers were 1.3\%, 5.0\%, and 6.7\%, respectively.
[Link to the Github repository:\\ \url{https://github.com/lnikolenko/EQA}]

\end{abstract}

\section{Introduction}
\subsection{Problem Statement}

Question Answering (QA) is essential to enabling effective communication between humans and machines. As a computer science discipline, it falls under information retrieval and natural language processing (NLP), and it concerns building machines able to answer questions asked by humans in natural languages \cite{cimiano2014ontology}. Recent years have seen significant progress in Question Answering owing to novel comprehensive public datasets, e.g. SQuAD \cite{rajpurkar2016squad}, TriviaQA \cite{joshi2017triviaqa}, HOTPOTQA \cite{yang2018hotpotqa}, and modern deep-learning models, most notably \textit{BERT} \cite{devlin2018bert}. As an instance of these advancements in Extractive Question Answering (EQA), state-of-the-art models trained on SQuAD dataset have surpassed human performance \cite{devlin2018bert}, while BERT model has a performance on par with human's on the updated version of the dataset, SQuAD 2.0 \cite{devlin2018bert}.

Yet, the mentioned achievements come at a price: \textit{massive human-made training datasets}. Generating this massive training data is typically crowd-sourced, and the process takes considerable time and resources \cite{lewis2019}. Further, training models on these large datasets is time-consuming and computationally expensive. The problem exacerbates when models are trained on languages other than English which is well-researched and has an abundance of training data available for different tasks. To tackle these challenges, some researchers have tried to create more effective models which can perform better than current models on existing datasets, while others have developed more complex datasets or devised methods of synthesizing training data to get better results from current EQA models. The following section briefly reviews some of these efforts during the past few years.

\subsection{Prior Research}
Researchers have adopted three major approaches for creating more effective QA systems: (1) create more effective models to better leverage existing datasets, (2) generate more actual training/test data using crowd-sourcing or (3) generate synthetic training/test data to improve the performance of existing models. These approaches are briefly explored below.

\subsubsection{Model Development}
Qi et al. \cite{qi2019answering} focus on the task of QA across multiple documents which requires multi-hop reasoning. Their main hypothesis is that the current QA models are too expensive to scale up efficiently to open-domain QA queries, so they create a new QA model called GOLDEN (Gold Entity) Retriever, able to perform iterative-reasoning-and-retrieval for open-domain multi-hop question answering. They train and test the proposed model on HOTPOTQA multi-hop dataset. One highlight of Qi et al. model is that they avoid computationally demanding neural models, such as BERT, and instead use off-the-shelf information retrieval systems to look for missing entities. They show that the proposed QA model outperforms several state-of-the-art QA models on HOTPOTQA test-set. 

In another work, Wang et al. \cite{wang2018r} develop an open-domain QA system, called $R^3$, with two innovative features in its question-answering pipeline: a ranker to rank the retrieved passages (based on the likelihood of retrieving the ground-truth answer to a query), and a reader to extract answers from the ranked passages using Reinforcement Learning (RL). Modern deep learning models for open-domain QA use large text corpora as training sets, and use a two-step process to answer questions: (1) Information Retrieval (IR) to select the relevant passages, and (2) Reading Comprehension (RC) to select candidate phrases containing the answer \cite{chen2017reading,dhingra2016gated}. The model proposed in this paper follows this same structure: Ranker module acts as the IR while Reader module acts as the RC. Wang et al. use SGD/backprop to train their Reader and to maximize the probability that the selected span contains the potential answer to the query. They train the Ranker using REINFORCE \cite{williams1992simple} RL algorithm with a reward function evaluating the quality of the answers extracted from the passages Ranker sends to Reader. They show that this configuration is robust against semantical differences between the question and the passage.

In an older study, Lee et al. \cite{lee2016learning} implemented a recurrent network (called RASOR) on SQuAD dataset for question answering, which resulted in a model with higher EM and F1 score compared to the most successful models up to the date [Match-LSTM]. In analysis, Lee et al. state that a recurrent net enables sharing computation for shared substructures across candidate spans for answering the asked question, and this has resulted in the improved performance of their model compared to the baseline models studied in the paper.

\subsubsection{Actual Data Generation}
Lewis et al. \cite{lewis2019mlqa} took on the challenge of `cross-lingual EQA' by developing a multi-lingual benchmark dataset, called \textit{MLQA}. This dataset covered seven languages including English and Vietnamese with more than 12k instances in English and 5k in the other six languages. They also managed to make each instance included in the benchmark to be paralleled across at least four of their chosen languages. Lewis et al. aimed to reduce the overfit observed in cross-lingual QA models. As their baseline models, Lewis et al. used BERT and XLM models \cite{lewis2019mlqa}. The dataset they developed only included development and testing set, so for training baseline models, they used the SQuAD v1.1 dataset. Using their test/dev dataset, Lewis et al. finally showed that the transfer results for state-of-the-art models (in terms of EM and F1 score) largely lag behind the training results; hence, more work is required to reduce the variance of high-performance models in EQA. 

In a recent paper, Reddy et al. \cite{reddy2019coqa} develop a dataset focused on Conversational Question-Answering, called CoQA. They hypothesize that machine QA systems should be able to answer questions asked based on conversations, as humans can do. Their dataset includes 127k question-answer pairs from 8k conversation passages across 7 distinct domains. Reddy et al. show that the state-of-the-art language models (including Augmented DrQA and DrQA+PGNet) are only able to secure an F1 score of 65.4\% on CoQA dataset, falling short of the human-performance by more than 20 points. The results of their work shows a huge potential for further research on conversational question answering which is key for natural human-machine communication. Previously, Choi et al. \cite{choi2018quac} had conducted a similar study on conversational question answering, and using high-performance language models, they obtained an F1 score 20 points less than that of humans on their proposed dataset, called QuAC.

In another work, Rajpurkar et al. \cite{rajpurkar2018know} focus on augmenting the existing QA datasets with unanswerable questions. They hypothesize that the existing QA models get trained only on answerable questions and easy-to-recognize unanswerable questions. To make QA models robust against unanswerable questions, they augment SQuAD dataset with 50k+ unanswerable questions generated through crowd-sourcing. They observe that the strongest existing language models struggle to achieve an F1 score of 66\% on their proposed update to SQuAD dataset (called SQuAD 2.0), while achieving an F1 score of 86\% on the initial version of the dataset. Rajpurkar et al. state that this newly developed dataset may spur research in QA on stronger models which are robust against unanswerable questions.

\subsubsection{Synthetic Data Generation}
Lewis et al. \cite{lewis2019} take on the challenge of expensive data-generation for Question Answering task by generating data and training QA models on synthetic datasets. They propose an unsupervised model for question-generating which powers the training process for an EQA model. Lewis et al. aim to make possible training effective EQA models with scarce or lacking training data, especially in non-English contexts. Their question-generation framework generates training data from Wikipedia excerpts. Training data in this work is generated as follows:
\begin{enumerate}
    \item A paragraph is sampled from English Wikipedia
    \item A set of candidate answers within that context get sampled using pre-trained models, such as Named-Entity Recognition (NER) or Noun-Chunkers, to identify such candidates
    \item Given a candidate answer and context, “fill-in-the-blank” cloze questions are extracted
    \item Cloze questions are converted into natural questions using an unsupervised cloze-to-natural-question translator.
\end{enumerate}
The generated data is then supplied to question-answering model as training data. BERT-LARGE model trained on this data can achieve 56.4\% F1 score, largely outperforming other unsupervised approaches. Before this paper, (i) generating training data for SQuAD question-answering and (ii) using unsupervised methods [instead of supervised methods] to generate training data directly on question-answering task were not explored as thoroughly. 

In a similar work, Zhu et al. \cite{zhu2019learning} propose a model to automatically generate unanswerable questions based on paragraph-answerable-question pairs for the task of machine reading comprehension. They use this model to augment SQuAD 2.0 dataset and achieve improved F1 scores, compared to the non-augmented dataset, using two state-of-the-art QA models. To create the model for generating unanswerable-questions, Zhu et al. adopt a pair-to-sequence architecture which they show outperforms models with a typical sequence-to-sequence question-generating architecture.

In an earlier work from 2017, Duan et al. \cite{duan2017question} propose a question-generator which can use two approaches for generating questions from a given passage (in particular, Community Question Answering websites): (1) a Convolutional Neural Network model for a retrieval-based approach, and (2) a Recurrent Neural Network model for a generation-based approach. They show that the questions synthesized by their model (based on data from YahooAnswers) can outperform the existing generation systems (based on BLEU metric), and it can augment several existing datasets, including SQuAD and WikiQA, for training better language models.

\subsection{Objective and Contributions} \label{obj}

This paper hypothesizes that for the task of Question Answering (QA), augmenting real data with synthesized data can help train models with a better performance compared to models trained only on real data. This work validates this on the task of Extractive Question Answering (EQA) using BERT language model \cite{devlin2018bert} trained on different combinations of real and artificial data, based on SQuAD 2.0 \cite{rajpurkar2018know} dataset (as the source of real data) and machine-generated answerable and unanswerable question-answer pairs (as the source of synthetic data). We will use F1 and Exact Match (EM) metrics to measure the performance of the developed models. We use an unsupervised generator-discriminator model based on cloze translation to generate answerable questions, following the work by Lewis et al. \cite{lewis2019}, and then alter the model to enable it to generate unanswerable questions. We expect the language model trained on augmented data to outperform the model trained on vanilla real data. We also expect models trained on synthetic data composed of both ANS and UNANS questions to yield better results than those trained on synthetic data composed of only ANS or only UNANS questions.

\section{Methodology}
\subsection{Model}
BERT model trained on 20\% of SQuAD 2.0 dataset will act as our baseline model. Improved models will be created by training BERT model on SQuAD 2.0 augmented with (1) answerable questions (ANS) from the work by Lewis et al. \cite{lewis2019}, (2) UNANS questions (UNANS) generated by the authors of this paper, and (3) a mixture of ANS and UNANS questions. Section \Ref{sec:exp} provides more details on the experiment designs. The following paragraphs describe the models used to generate the ANS and UNANS datasets.

The model generating synthetic answerable questions was developed by Lewis et al. \cite{lewis2019}. It takes as its input a paragraph from English Wikipedia, and uses a Named Entity Recoginition (NER) system to identify a set of potential answers which it then uniformly samples from. Next, an answer $a$ is generated by identifying a sub-clause around the named entity using an English syntactic parser. To generate the maximum likelihood question $p(q | a, c)$ from the context $c$ (the paragraph) and answer $a$, the model produces a cloze statement — i.e.\ a statement with a masked answer — from the identified sub-clause. An example would be ``I ate at McDonald's" which maps to ``I ate at [MASK]". Then the system uses unsupervised Neural Machine Translation (NMT) \cite{lample2018phrase} to translate the cloze question into a natural question, and it finally outputs the generated question-answer pair.

We plan to enhance Lewis et al.'s model by enabling it to generate both ANS and UNANS questions. To do this, we will first refactor the model so that instead of treating each paragraph as a standalone article, it can generate question-answer pairs for multiple paragraphs within a single article. Next, to make it possible to use SQuAD 2.0 as our training set, we will modify the model to accept inputs and produce outputs with the standard format of the SQuAD 2.0 dataset. To produce UNANS questions, we will shuffle the questions about the input paragraphs within the same article: This ensures that the questions are indeed unanswerable, since they will be detached from their original context, while staying relevant to the original paragraphs. Sustaining this relevance also helps make the unanswerable questions resilient against word-overlap heuristic \cite{yih-etal-2013-question} because the paragraphs will belong to the same article. 

At the end, we will evaluate how well the synthetic training examples complement the SQuAD 2.0 human-labeled data: We will use EM and F1 scores to assess the performance of BERT model (implemented by HuggingFace\footnote{\url{https://huggingface.co/}}) on EQA among models trained only on human-generated data and models trained on human-generated data combined with the two sets of synthetic datasets, i.e.\ ANS and UNANS examples.

\subsection{Data}

Here, we train the language models for EQA on the renowned Stanford Question Answering Dataset (SQuAD) 2.0 \cite{rajpurkar2018know}. This dataset is an updated version of SQuAD 1.0 \cite{rajpurkar2016squad} which was a reading comprehension dataset comprised of 100k+ questions built around Wikipedia articles. SQuAD 2.0 was created by adding 50k crowd-sourced (adversarial) unanswerable questions to the initial dataset.

\begin{figure}[h]
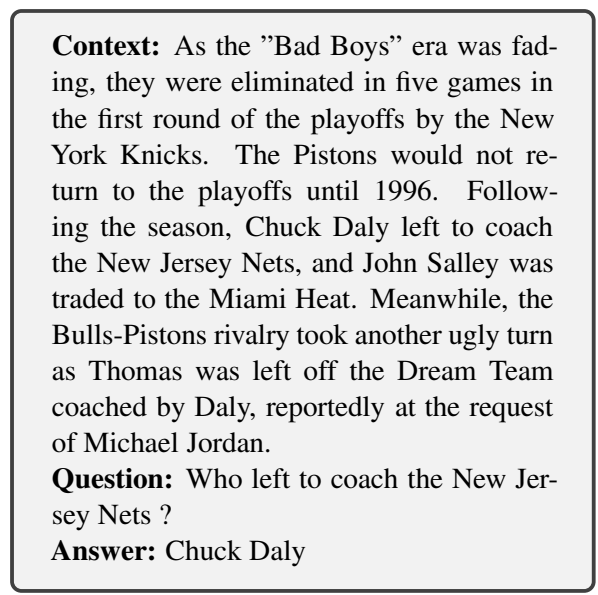


\begin{tcolorbox}

\textbf{Context:} As the "Bad Boys" era was fading, they were eliminated in five games in the first round of the playoffs by the New York Knicks. The Pistons would not return to the playoffs until 1996. Following the season, Chuck Daly left to coach the New Jersey Nets, and John Salley was traded to the Miami Heat. Meanwhile, the Bulls-Pistons rivalry took another ugly turn as Thomas was left off the Dream Team coached by Daly, reportedly at the request of Michael Jordan.

\textbf{Question:} Who left to coach the New Jersey Nets ?

\textbf{Answer:} Chuck Daly
\end{tcolorbox}
\caption{An example of a synthetic answerable question-answer pair.}
\label{fig:lewis_example}
\end{figure}

\begin{figure}[h]
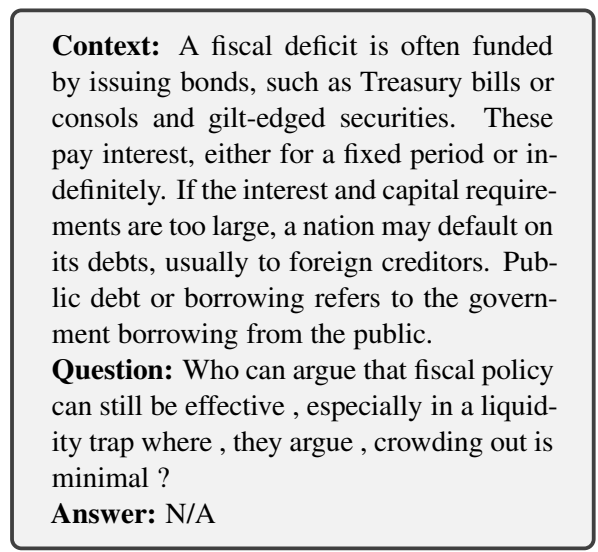


\begin{tcolorbox}

\textbf{Context:} A fiscal deficit is often funded by issuing bonds, such as Treasury bills or consols and gilt-edged securities. These pay interest, either for a fixed period or indefinitely. If the interest and capital requirements are too large, a nation may default on its debts, usually to foreign creditors. Public debt or borrowing refers to the government borrowing from the public.

\textbf{Question:} Who can argue that fiscal policy can still be effective , especially in a liquidity trap where , they argue , crowding out is minimal ?

\textbf{Answer:} N/A
\end{tcolorbox}
\caption{An example of a synthetic unanswerable question-answer pair.}
\label{fig:unanswerable_example}
\end{figure}

As the source of answerable synthetic questions, we use the dataset generated by Lewis et al. \cite{lewis2019}\footnote{\url{https://github.com/facebookresearch/UnsupervisedQA}}. See figure \ref{fig:lewis_example} for a synthetic question-answer example. The dataset contains ~3.9M answerable question-answer pairs created using a cloze-translating generator. This data is generated in SQuAD 1.0 standard format: we will convert the data into SQuAD 2.0 format to be able to merge it with human-generated question-answer pairs from SQuAD 2.0 dataset. The dataset Lewis et al. \cite{lewis2019} generated with their model includes \textit{only} answerable questions. To generate the required unanswerable data, we modify their data-generation pipeline. We have used pre-processed Wikipedia dump\footnote{\url{https://dumps.wikimedia.org/}} as an input to the updated/modified question-answer generation model to generate around 80k unanswerable training examples in SQuAD 2.0 format. Figure \ref{fig:unanswerable_example} contains an instance of the generated unanswerable question-answer pair.  

\subsection{Metrics}
We will use macro-averaged F1 score and EM to evaluate the performance of the models trained in this work. F1 score shows the precision and recall for the words selected as part of the answer actually being part of the correct answer. We first compute the F1 score of the model's predictions against the ground-truth answer represented as bags of tokens, then take the maximum F1 score across all possible answers for a given question, and finally average over all of the questions. EM, on the other hand, indicates the number of exactly correct answers with the same start and end indices.

\section{Experiments} \label{sec:exp}
We expect to achieve more reliable models for the task of EQA when augmenting actual training data with synthetic data. The synthetic data (question-answers) used for augmenting the actual data in this project has two types: answerable and unanswerable questions. Lewis et al. \cite{lewis2019} observed that synthetic answerable questions can boost the performance of QA models when added to actual data from SQuAD 1.1 dataset. Also, Zhu et al. \cite{zhu2019learning} observed that mixing synthetic unanswerable questions derived from human-generated training examples into actual data from SQuAD 2.0 can improve the performance of EQA models. We hence expect that augmenting an actual dataset, \textit{i.e.} SQuAD 2.0 in this work, with a mix of ANS and UNANS can yield an even better performance than using each of them alone to enhance the dataset.

We have devised several experiments to test the mentioned hypothesis with training examples described below:
\begin{enumerate}
    \item Experiment 0 [Baseline]: Using 26,063 examples from SQuAD 2.0 dataset [the entire dataset was not selected to make the training tractable]
    \item Experiment 1-1 [ANS Augmentation]: Using 26,063 examples from SQuAD 2.0, and 391,549 from ANS (from \cite{lewis2019})
    \item Experiment 1-2 [UNANS Augmentation]: Using 26,063 examples from SQuAD 2.0, and 76,818 from UNANS
    \item Experiment 2 [ANS+UNANS Augmentation]: Using 26,063 examples from SQuAD 2.0, 314,731 from ANS, and 76,818 from UNANS
\end{enumerate}
Experiment 0 provides a baseline to compare the other experiment results against. Experiment 1 looks into the impact of exclusive ANS or UNANS data augmentation. Finally, Experiment 2 will show the results of mixing the two approaches of augmentation together. 

BERT model adapted to EQA was used to run the mentioned experiments, and the results were evaluated on a set of held-out human-generated data points consisting of 3,618 question-answer pairs. We have tuned the hyper-parameters of the model (number of training epochs, maximum sequence length, etc.) based on our observations from Experiment 0, since it involves a relatively small dataset and is easy to experiment with. We will use these obtained optimal hyper-parameters for the rest of the experiments. During our initial experimentation, we observed that training BERT on the full SQuAD 2.0 dataset takes 9 hours on a 1480 MHz 3584 core NVIDIA 1080 TI GPU, so to avoid excessive training times, we decided to use only 20\% of the SQuAD dataset and accordingly use a limited portion of the synthetic questions generated by Lewis et al. \cite{lewis2019}.

\begin{table}[]
    \centering
    \begin{tabular}{||c c c||}
     \hline
    Experiment & F1 (\%) & EM (\%)\\ [0.5ex] 
     \hline\hline
     0 & 57.61 & 61.27 \\ 
     \hline
     1-1 & 58.90 & 62.56\\ [1ex] 
     \hline
     1-2 & 62.56 & 65.81 \\ [1ex] 
     \hline
     2 & 64.28 & 66.36 \\ [1ex] 
     \hline
    \end{tabular}
    \caption{Results of the three experiments}
    \label{tab:1}
\end{table}

\section{Results and Discussion}
Table \ref{tab:1} shows the results of the experiments. A few observations can be made based on these results:

\begin{itemize}
    \item Experiments 1-1 and 1-2 demonstrate that, as expected, adding either ANS or UNANS questions to the human-generated training examples boosts F1 and EM scores of the BERT model for both cases compared to Baseline.
    \item The results further show that adding the ANS data to the original dataset (experiment 1-1) has a stronger impact than adding the UNANS data (experiment 1-2). Table \ref{tab:2} indicates this point: the normalized impact of adding a single example from the UNANS dataset is almost four-times larger than that of the ANS dataset on the F1 and EM scores compared to the baseline. This can be justified with the following: the original training set has a small portion (only around 1/3) of unanswerable questions, so our synthetic dataset increases the proportion of unanswerable questions and makes the training data more balanced in this regard.
    \item Finally, the results of experiment 2 show that augmenting the SQUAD 2.0 dataset with both ANS and UNANS at the same time leads to an even greater performance compared to using either of the two datasets to enhance the human-made data, \textit{i.e.} compared to experiments 1-1 and 1-2. 
\end{itemize}
These results confirm our hypothesis mentioned in section \ref{obj}, and show a potential for our novel synthesized unanswerable dataset to further boost the performance of language models similar to BERT for the task of EQA. 

\begin{table}[]
    \centering
    \begin{tabular}{||c c c||}
     \hline
     & ANS & UNANS\\ 
     \hline\hline
     Gain in F1 (\%/example) & 0.022 & 0.086 \\ 
     \hline
     Gain in EM (\%/example) & 0.021 & 0.074\\ 
     \hline
    \end{tabular}
    \caption{Comparison between the relative impact of each dataset on the model scores: ANS vs.\ UNANS}
    \label{tab:2}
\end{table}




\section{Conclusions}
This paper studies the impact of augmenting human-made data with synthetic data on the task of Extractive Question Answering by using BERT \cite{devlin2018bert} as the language model and SQuAD 2.0 \cite{rajpurkar2018know} as the baseline dataset. Two sets of synthetic data are used for augmenting the baseline data: a set of answerable and another set of unanswerable questions-answers. Conducted experiments show that using both these synthetic datasets can tangibly improve the performance of the selected language model for EQA, while the UNANS data, generated by the authors, has a more pronounced impact on improving the performance. Adding the UNANS dataset to the original data yields a gain of ~5\% in both F1 and EM scores, whereas the ANS dataset yields around a quarter of this gain. Enhancing the original data with a combination of the two synthetic datasets improves the F1 score of BERT on the test-set by ~7\% and the EM score by ~5\% which are sizable improvements compared to the performance of the baseline models and similar efforts in the literature. The obtained results indicate the great potential of using synthetic data to complement the costly human-generated datasets: This augmentation can help provide the massive data required for training the modern language models at a very low cost.

\section{Limitations}
The presented approach has limitations similar to \cite{lewis2019}: Although we tried to avoid using any human-labeled data for generating the synthetic question-answers, the question-generating models rely on manually-labeled data from OntoNotes 5 (for NER system) and Penn Treebank (for extracting subclauses). Further, the question-generation pipeline of this work uses English language-specific heuristics. Hence, the applicability of this approach is limited to languages and domains that already have a certain amount of human-labeled data for question generation, and porting this model to another language would require extra preparatory efforts. 

An extensive amount of training examples are required to achieve tangible performance gains, and this results in substantial training times and compute costs for both generating synthetic data and training the BERT model. These high training times and resource costs prevented us from performing the experiments on the full SQuAD 2.0 dataset. Nonetheless, given the homogeneity of the original dataset, we expect the synthetic training examples to bring similar performance improvements if added to the full dataset with similar proportions.

\section{Future Work}
The work presented in this manuscript can be extended in several ways:
\begin{itemize}
    \item Developing a more sophisticated unsupervised model for unanswerable question generation can be a great extension of this work. Some potential approaches include coming up with heuristics such as word/synonym overlap for filtering the generated questions and employing the pair-to-sequence model by Zhu et al. \cite{zhu2019learning} on the synthetic training data.
    \item The computational power available to the authors limited the size of the data used for running the experiments in this work: future efforts can run more extensive experiments to further examine the synthetic data augmentation studied here.
    \item Breaking down the question types into how, what, where, when, etc. and studying the individual impacts of each question-answer type can also shed more light on the individual impact of each question type on the performance of the language model. The insights gained from such experiment can help fine-tune the generated data to achieve more effective synthetic datasets.
\end{itemize}



\section{Acknowledgments}
We would like to thank Stanford's CS224N and CS224U course staff, especially Professor Chris Potts, for their guidance and feedback on this project. 
\section{Authorship Statements}
Liubov implemented unanswerable question generation pipeline and the scripts to process and partition the data. Pouya worked on designing the experiments and composing the paper.

\bibliographystyle{acl_natbib}
\bibliography{emnlp2020}

\end{document}